\documentclass[11pt]{article}
\usepackage{coling2018}
\usepackage{times}
\usepackage{url}
\usepackage{latexsym}

\usepackage{array}
\usepackage{url}
\usepackage{caption}
\usepackage{chngpage}
\usepackage{subfigure}
\usepackage{graphicx}
\usepackage{amsfonts}
\usepackage{amsmath}
\usepackage{CJK}
\usepackage{multirow}
\usepackage{xcolor}

\title{Fusing Recency into Neural Machine Translation \\with an Inter-Sentence Gate Model}

\author{
Shaohui Kuang\hspace{0.5cm} Deyi Xiong\thanks{ \hspace{0.1cm} Corresponding author} \\
School of Computer Science and Technology, Soochow University, Suzhou, China \\
{\tt shaohuikuang@foxmail.com, dyxiong@suda.edu.cn}
}

\date{}

\begin{document}
\begin{CJK}{UTF8}{gbsn}
\maketitle
\begin{abstract}
Neural machine translation (NMT) systems are usually trained on a large amount of bilingual sentence pairs and translate one sentence at a time, ignoring inter-sentence information. This may make the translation of a sentence ambiguous or even inconsistent with the translations of neighboring sentences. In order to handle this issue, we propose an inter-sentence gate model that uses the same encoder to encode two adjacent sentences and controls the amount of information flowing from the preceding sentence to the translation of the current sentence with an inter-sentence gate. In this way, our proposed model can capture the connection between sentences and fuse recency from neighboring sentences into neural machine translation. On several NIST Chinese-English translation tasks, our experiments demonstrate that the proposed inter-sentence gate model achieves substantial improvements over the baseline.
\end{abstract}

\section{Introduction}

    %
    %
    
    %
    %
    %
    %

In NMT systems \cite{bahdanau2015neural,cho2014properties,sutskever2014sequence}, an encoder first reads variable-length source sentences and encodes them into a sequence of vectors, then a decoder generates a target translation from the sequence of source vectors. Although NMT is an emerging machine translation approach, the translation process of a document in NMT is similar to conventional statistical machine translation (SMT), treating the document as a bag of sentences and ignoring cross-sentence dependencies.

Sentences are the constituent elements of paragraphs. \newcite{harper1965studies} argues that sentences possess two attributes: continuity that maintains consistency with other sentences and development that introduces new information. For any adjacent sentences of a well-formed text, they tend to have considerable degree of continuity, which is usually described and measured at two levels: cohesion at the surface level and coherence at the underlying level. These two continuity metrics are two well-known means to estalish such inter-sentence links within a text. \newcite{harper1965studies} finds that word recurrence, as the most common device of cohesion,  occurs in 70\% of adjacent sentence pairs. \newcite{Xiong2015Topic} show that about 60\% of sentences have the same topics (coherence) as those of the documents where these sentences occur. 

Such inter-sentence dependencies can and should be used to help document-level machine translation. In the literature, a variety of models have been proposed to capture these dependencies in the context of SMT, such as cache-based language and translation models \cite{tiedemann2010context,gong2011cache}, topic-based coherence model \cite{xiong2013topic} and lexical cohesion model \cite{xiong2013modeling}. However, integrating inter-sentence information into an  NMT system is still an open problem. 

In this paper, we propose a simple yet effective approach to model the inter-sentence information for NMT. In order to capture  the connection between two adjacent sentences, i.e. the preceding sentence and current sentence, we first use the same encoder to encode the two adjacent sentences at the same time to form a context vector {\it a} for the preceding sentence and a context vector {\it b} for current sentence. Then, we introduce a gate mechanism to combine {\it a} and {\it b} into the final context vector, which is further used to update the hidden states of the decoder. In this way, we can model the links and dependencies between adjacent sentences. To some extent, our approach models the inter-sentence relationship from the underlying semantic coherence perspective.

On the NIST Chinese-English translation tasks, our experimental results show that the proposed approach can achieve significant improvements of up to 2.0 BLEU points over the NMT baseline. 

\section{Related Work}

In the literature, a series of document-level translation models have been proposed for conventional SMT. Just to name a few, \newcite{gong2011cache} propose a cache-based approach to document-level translation, which includes three caches, a dynamic cache, a static cache and a topic cache to capture various kind of document-level information. 
\newcite{hardmeier2012document} present a beam search decoding procedure for phrase-based SMT with features modeling cross-sentence dependencies.
\newcite{xiong2013topic} propose a topic-based coherence model to produce discourse coherence for document translation. \newcite{xiong2013modeling} present a lexical cohesion model to capture lexical cohesion for document-level translation. 

In neural language models, inter-sentence connections can be captured in a contextual model. For example, \newcite{Lin2015Hierarchical} propose a hierarchical recurrent neural network (HRNN) language model for document modeling, consisting of a sentence-level and word-level language model, and use the proposed model to model sentence-level coherence. In speech recognition, as input speech signals can contain thousands of frames, \newcite{chan2016listen} employ Bidirectional Long Short Term Memory with a pyramidal structure to capture the context of a large number of input time steps. \newcite{Wang2016LARGER} introduce a late fusion method to incorporate corpus-level discourse information into recurrent language modeling.

In neural conversation systems, links between multi-turn conversations are usually modeled with hierarchical neural networks. 
\newcite{Serban2015Building} use a hierarchical recurrent encoder-decoder(HRED) to model the dialogue into two-level hierarchy: a sequence of utterances and a sequence of words. The proposed model can track states over many utterances to generate context-aware multiple rounds of dialogue. 
\newcite{Serban2016A} further propose a HRED model with an additional component: a high-dimensional stochastic latent variable at every dialogue turn to sample a vaussian variable as input to the decoder.

It is natural to adapt the HRED model to document-level NMT. However,  document boundaries are usually missing in bilingual training corpora, indicating that we do not have sufficient data to train the sentence-level hierarchy. In our proposed gate model, we do not need entire documents to train the NMT model. We only use pairs of two adjacent sentences to train the gate. Furthermore, each sentence in a sentence pair can be used twice: one as a preceding sentence and the other as a current sentence. Additionally, in order to reduce the number of extra parameters, we use the same encoder to encode the adjacent sentences, which can also keep the semantic consistency for the same source sentence.

\section{Neural Machine Translation}
In this section, we briefly describle the atttention-based NMT model proposed in \cite{bahdanau2015neural}.

In their framework, the encoder encodes a source sentence as a sequence of vectors with bi-directional RNNs. The forward RNN reads the source sentence \(x = (x_1, x_2, ..., x_T)\) from left to right and the backward RNN reads the source sentence in an inverse direction. The hidden states \(\overrightarrow{h} = (\overrightarrow{h_1}, \overrightarrow{h_2}, ..., \overrightarrow{h_T})\)  in the forward RNN can be computed as follows:
\begin{equation}
\overrightarrow{h_j} = f(\overrightarrow{h_{j-1}}, x_j),
\end{equation}
where \(f\) is a non-linear function, here defined as a gated recurrent unit (GRU) \cite{chung2014empirical}. Similarly, Hidden states of the backward RNN can be calculated. The forward and backward hidden states are concatenated into the final annotation vectors \(h =(h_1, h_2, ..., h_T)\). 
The decoder is also an RNN that predicts the next word \(y_t\) given the context vector \(c_t\)
, the hidden state \(s_{t-1}\) and the previously generated word sequence \(y_{<t} = [y_1, y_2, ..., y_{t-1}]\). The probability of the next word \(y_t\) is calculated as follows:
\begin{equation}
p(y_t|y_{<t};x) = g(c_t, y_{t-1}, s_t),
\end{equation}
where \(g\) is a softmax layer, \(s_t\) is the state of decoder RNN at time step \(t\) computed as 
\begin{equation}
s_t = f(s_{t-1}, y_{t-1}, c_t).
\end{equation}
where \(f\) is a function, the same as function used in the encoder. The context vector \(c_t\) is calculated as a weighted sum of all hidden states of the encoder as follows:
\begin{equation}
c_t = \sum_{i=1}^{T_x} \alpha_{tj}h_j,
\end{equation}
\begin{equation}
\alpha_{tj} = \frac{exp(e_{tj})}{\sum_{k=1}^{T_x} exp(e_{tk})},
\end{equation}
\begin{equation}
e_{tj} = a(s_{t-1},h_j).
\end{equation}
where \(\alpha_{tj}\) is the weight of each hidden state \(h_j\) computed by the attention model, \(a\) is a feedforward neural network with a single hidden layer.

\begin{figure}[!t]
\centering
\includegraphics[height=1.9in,width=2.8in]{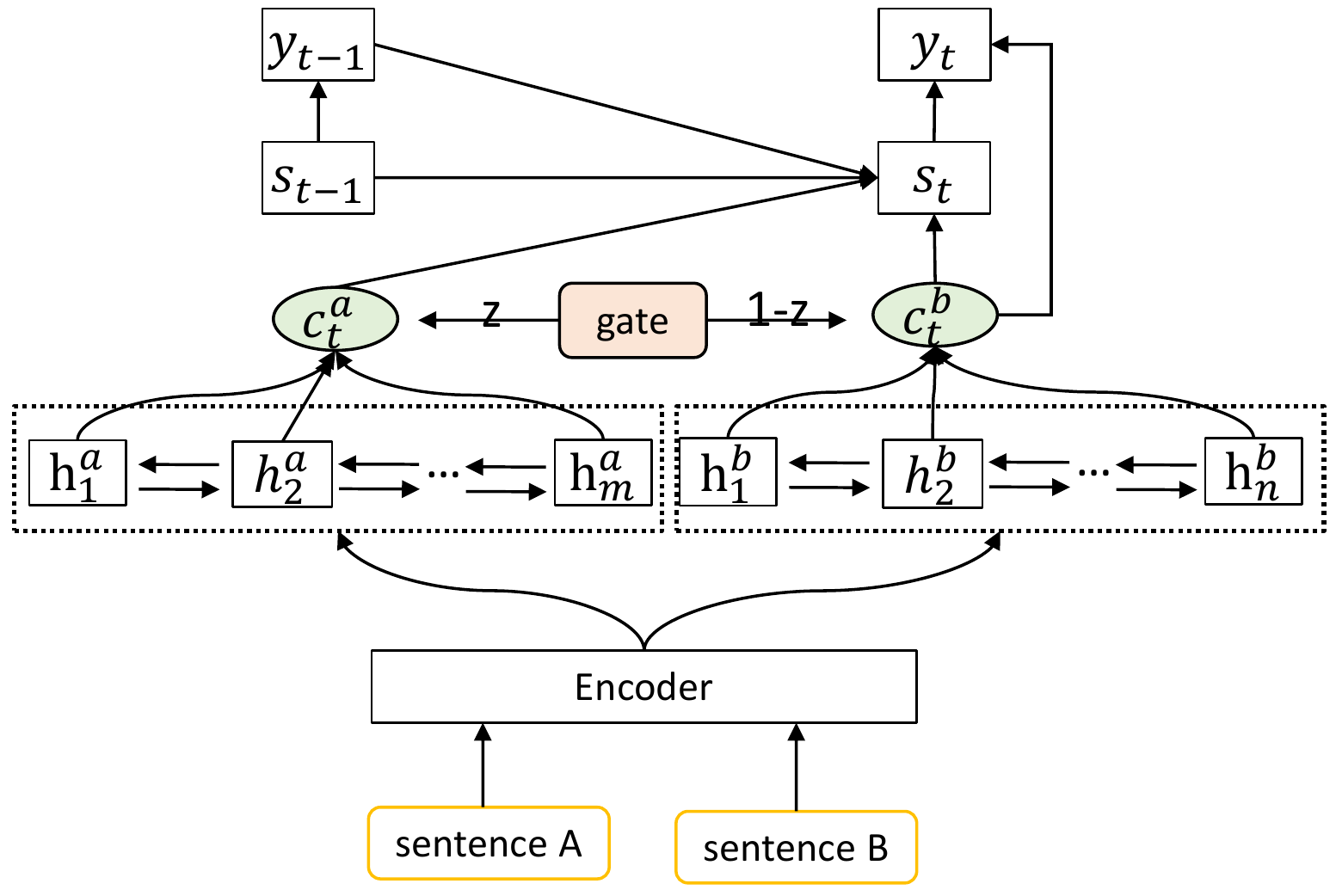}
\caption{Architecture of NMT with the inter-sentence gate.}
\label{fig:1}
\end{figure}

We also implement an NMT system which adopts feedback attention \cite{wang2016neural,wang2016memory}, which will be referred to as RNNSearch in this paper. In the feedback attention, \(e_{tj}\) is computed as follows:
\begin{equation}
e_{tj} = a(\widetilde s_{t-1},h_j),
\end{equation}
where \(\widetilde s_{t-1} = GRU(s_{t-1}, y_{t-1})\). The hidden state of the decoder is updated as follows:
\begin{equation}
s_t = GRU(\widetilde s_{t-1}, c_t)
\end{equation}

In this paper, our proposed model is implemented on the top of RNNSearch system.

\section{The Inter-Sentence Gate Model}
 
In this section, we will elaborate the proposed inter-sentence gate model, which we refer to as \(NMT_{ISG}\). Figure 1 shows the entire architecture of our NMT with the inter-sentence gate. For notational convenience, we denote two adjacent sentences as A and B: A for the preceding sentence and B for the current sentence.

\subsection{Encoder}
We employ the same encoder to encode the adjacent sentence A and B into hidden vector representations \([h_1^a, h_2^a, ......, h_m^a]\) and \([h_1^b, h_2^b, ......, h_n^b]\) respectively. We then use the attention network described in Equation (4) of Section 3 to compute their context representations \(c_t^a\) and \(c_t^b\).

\subsection{Inter-Sentence Gate Model}

When we translate the current sentence B, we have to make sure that the decoder is provided with sufficient information from sentence B, and at the same time, with helpful imformation from the preceding sentence A. In other words, we need a mechanism to control the scale of information flowing from the sentence A and sentence B to the decoder.
Inspired by the success of gated units in RNN \cite{chung2014empirical}, we propose an inter-sentence gate to control the amount of information flowing from A and B. Formally, we employ a sigmoid neural network layer and an element-wise multiplication operation, as illustrated in Figure 2.  Similarly, \newcite{tu2016context} also propose a gating mechanism to combine source and target contexts. The gate framework assigns element-wise weights \(z\) to the input signals, calculated by 
\begin{equation}
z_t = \sigma(U_zs_{t-1}+W_zy_{t-1}+C_bc_t^b+C_ac_t^a)
\end{equation}
here $\sigma$ is a logistic sigmoid function, and \(U_z\), \(W_z\), \(C_b, C_a\) are the parameter matrix.

\begin{figure}[!t]
\centering
\includegraphics[height=1.4in,width=2.4in]{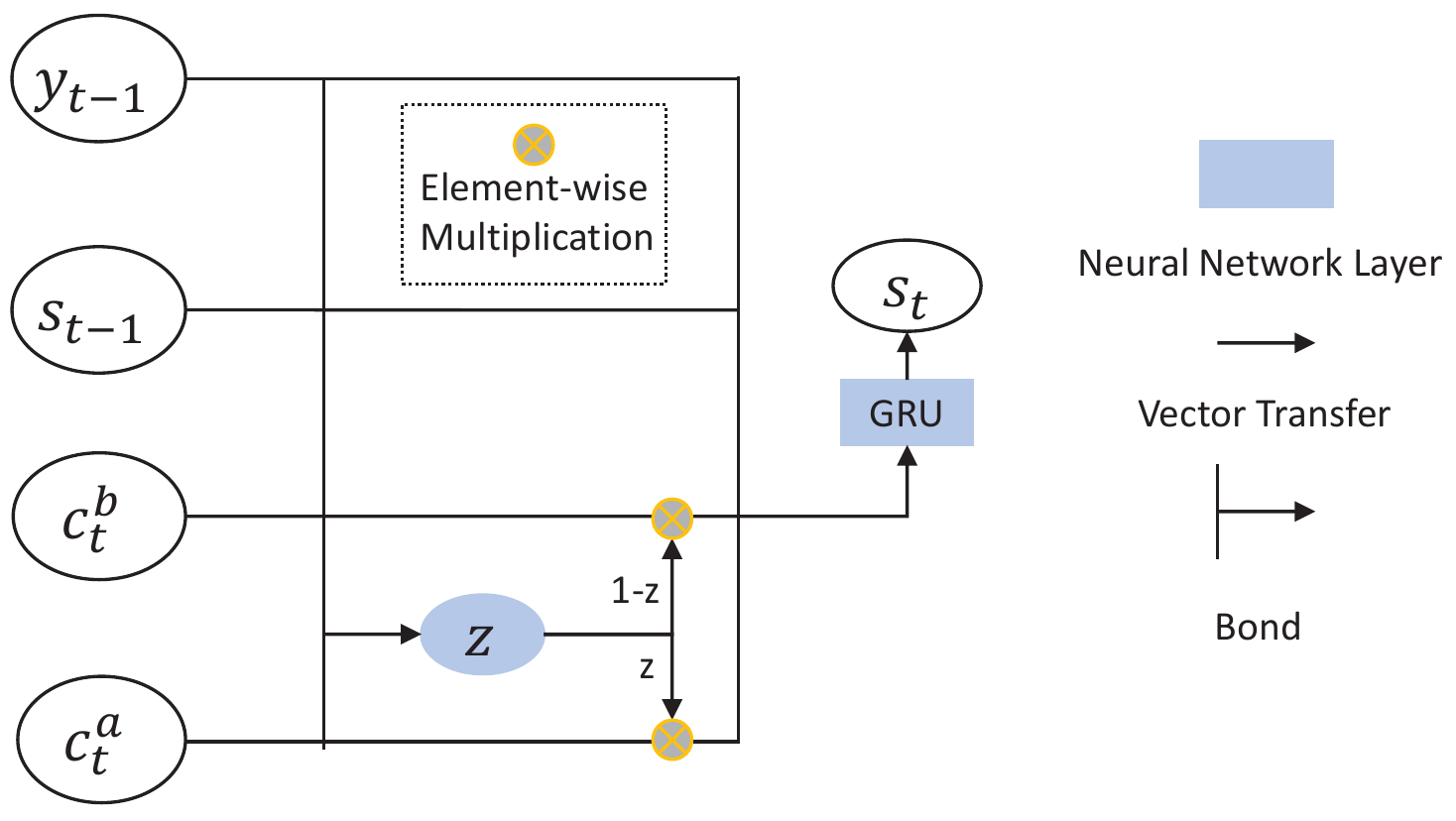}
\caption{The framework of updating the decoding state using the proposed gate model.} 
\label{fig:3}
\end{figure}

\subsection{Decoder}

Next, we integrate the inter-sentence gate into the decoder to decide the amount of context information used in producing the decoder hidden state at each time step. In this way, we want the hidden states of the decoder to store the inter-sentence context information. 
The framework of updating the hidden state of the decoder at time step \(t\) is illustrated in Figure 2.
The decoder hidden state $s_t$ is computed as follows:
\begin{equation}
\begin{split}
&s_t = GRU(Us_{t-1}+Wy_{t-1}+C_1c_t^a*z_t+C_2c_t^b*(1-z_t)) 
\end{split}
\end{equation}
where \(*\) is an element-wise multiplication, \(U, W, C_1, C_2\) is the parameter matrix, and $z_t$ is the inter-sentence gate computed by Equation (10). 

The conditional probability of the next word \(y_t\) is calculated as follows:
\begin{equation}
p(y_t|y_{<t},x) = g(f(s_t,y_{t-1}, c_t^b))
\end{equation}
where \(c_t^b\) is the context vector of the current sentence B.

Our aim is to translate the current sentence B with the additional information from the preceding sentence A. We do not want to have the excessive impact of the preceding sentence on the translation output of the current sentence. Therefore, in the stage of generating the next word, we just use the context \(c_t^b\).

\section{Experiments}

We carried out a series of Chinese-to-English translation experiments to evaluate the effectiveness of the proposed inter-sentence gate model on document-level NMT and conducted in-depth analyses on experiment results and translations.

\subsection{Experimental Settings}
We selected corpora LDC2003E14, LDC2004T07, LDC2005T06, and LDC2005T10 as our bilingual training data, where document boundaries are kept. We also used all data from the corpus LDC2004T08 (Hong Kong Hansards/Laws/News). In total, our training data contain 103,236 documents and 2.80M sentences. Averagely, each document consists of 28.4 sentences. We chose NIST05 dataset as our development set, and NIST02, NIST03, NSIT04, NIST06, NIST08 as our test sets. We used case-insensitive BLEU-4 as our evaluation metric. We compared our \(NMT_{ISG}\) with the following two systems:

\begin{itemize}
\item {\bf Moses} \cite{koehn2007moses}: an open phrase-based translation system with its default setting.
\item {\bf RNNSearch}: our new implementation of NMT system with the feedback attention as described in Section 3.
\end{itemize}

For Moses, we used the full training data (parallel corpus) to train the model. Word alignments were produced by GIZA++ \cite{och2000improved}. We ran GIZA++ on the corpus in both directions, and merged alignments in two directions with  ``grow-diag-final'' refinement rule \cite{koehn2005edinburgh}. We trained a 5-gram language model on the Xinhua portion of the Gigaword corpus using SRILM Toolkit with a modified Kneser-Ney smoothing.

For RNNSearch, we used the parallel corpus to train the attention-based NMT model. The encoder of RNNSearch consists of a forward (1000 hidden unit) and backward (1000 hidden unit) recurrent neural network. The maximum length of sentences that we used to train NMT in our experiments was set to 50 for both the Chinese and English sides. We used the most frequent 30K words for both Chinese and English, covering approximately 99.0\% and 99.2\% of the data in the two languages respectively. We replaced rare words with a special token “UNK”.  We also adopted the dropout technique. Dropout is applied only on the output layer and the dropout rate was set to 0.5. 
All the other settings are the same as the setting up described by \newcite{bahdanau2015neural}. Once the NMT model was trained, we employed a beam search algorithm to find possible translations with high probabilities. We set the beam width to 10. 


For the proposed \(NMT_{ISG}\) model, we implemented it on the top of RNNSearch. We used tuples (x, before-x, y) as input of \(NMT_{ISG}\), where x and y are a parallel sentence pair, before-x is the previous sentence of source sentence x in the same document\footnote{ We obtain tuples {(x, before-x, y)} from the training corpus. During the acquisition of these tuples, we follow two constraints. First, we discard the tuple (x,before-x,y) if there is a big difference in the length of the sentence x and before-x. For example, sentence  x or before-x is a date expression at the end of a document or an organization name at the beginning of a document. Second, the length of any element in a tuple (x, before-x, y) is not longer than 50. 
}. 
As the first sentence of a document does not have before-x, we used the stop symbol to form the sentence s = (eos, eos, eos) as the before-x. We used a simple pre-training strategy to train our \(NMT_{ISG}\) model: 
training the regular attention-based NMT model using our implementation of RNNSearch, and then using its parameters to initialize the parameters of the proposed model, except for those related to the operations of the inter-sentence gate. 

We used the stochastic gradient descent algorithm with mini-batch and Adadelta \cite{zeiler2012adadelta} to train the NMT model. The mini-batch was set to 80 sentences and decay rates $\rho$ and $\epsilon$ of Adadelta were set to 0.95 and \(10^{-6}\), respectively. 

\begin{table*}[!t]
\centering
\begin{tabular}{c|c|cccccc}
\hline 
\bf Model & \bf NIST05 & \bf NIST02 & \bf NIST03 & \bf NIST04 & \bf NIST06 & \bf NIST08 & \bf Avg\\ 
\hline
\hline
Moses              & 29.52 & 31.52  & 31.68 & 32.73 & 29.57 & 23.09 & 29.72 \\
\hline
\hline
RNNSearch          & 32.56 & 36.18 & 34.85 & 36.36 & 30.57 & 23.69 & 32.37 \\
\hline
\(NMT_{ISG}\)      & \(34.58^{\ddag}\) & \(36.68^{\ddag}\) & \(36.29^{\ddag}\) & \(38.15^{\ddag}\) & \(31.83^{\ddag}\) & \(24.67^{\ddag}\) & 33.7 \\
\hline
\end{tabular}
\caption{\label{font-table} Experiment results on the NIST Chinese-English translation task. We adopted the RNNSearch, an in-house NMT system, as our baselines. \(NMT_{ISG}\) is the proposed model without replacing UNK words. The BLEU scores are
case-insensitive. Avg means the average BLEU score on all the test sets. “\ddag”: statistically better than RNNSearch (p $<$0.01).}
\end{table*}

\begin{table*}[!t]
\centering
\begin{tabular}{c|c|cccccc}
\hline 
\bf Model  & \bf NIST02 & \bf NIST03 & \bf NIST04 & \bf NIST05  & \bf NIST06 & \bf NIST08 \\ 
\hline
\hline
RNNSearch                & 32.56 & 36.18 & 34.85 & 36.36 & 30.57 & 23.69 \\
\hline
RNNSearch + concat        & 21.81 & 18.72 & 19.44 & 16.11 & 16.60 & 9.81 \\

\hline                  
\end{tabular}
\caption{\label{font-table2} The BLEU scores of RNNSearch + concat model which uses the concatenation of two neighboring sentence as input of RNNSearch.}
\end{table*}

\subsection{Experimental Results}

Table 1 shows the results of different NMT systems measured in terms of BLEU score. From the table, we can find that our implementation RNNSearch using the feedback attention and dropout outperforms Moses by 2.65 BLEU points. The proposed model \(NMT_{ISG}\) achieves an average gain of 1.33 BLEU points over RNNSearch on all test sets. And it outperforms Moses by 3.98 BLEU points.

One might use the concatenation of two  neighboring source sentences as input of RNNSearch to explore the information of the preceding sentence. However, this will degenerate translation quality as shown in Table 2. The main reason is that the conventional NMT has difficulties in translating long sentences \cite{Pougetabadie2014Overcoming}. Thus，we conclude that the information of preceding sentences cannot be directly explored via concatenation. 



\begin{table*}[!t]
\centering
\begin{tabular}{c|c|cccccc}
\hline 
\bf Model & \bf NIST05 & \bf NIST02 & \bf NIST03 & \bf NIST04 & \bf NIST06 & \bf NIST08 & \bf Avg\\ 
\hline
\hline
RNNSearch                & 32.56 & 36.18 & 34.85 & 36.36 & 30.57 & 23.69 & 32.37 \\
\hline
\hline
\(NMT_{ISG}\)            & 34.58 & 36.68 & 36.29 & 38.15 & 31.83 & 24.67 & 33.7 \\
\hline                  
+NULL     & 33.22 & 36.46 & 35.27 & 37.11 & 30.77 & 23.80 & 32.77 \\ 
\hline
+$z$={\bf 0} & 30.49 & 31.71 & 31.05 & 35.10 & 29.89 & 22.10 & 30.06\\
\hline
+RV        & 31.88 & 36.21 & 34.51 & 35.93 & 29.96 & 22.99 & 31.91\\
\hline
\end{tabular}
\caption{\label{font-table3} Effect of the inter-sentence gate and information of before-x. BLEU scores in the table are case-insensitive. [+NULL] is set the before-x to a NULL sentence. [+$z$=0] is set the gate vector to all-zero vector. And [+RV] is set the context vector of before-x to a random vector which value of vector is between -1 and 1. Avg meaning the average BLEU score on all the test sets.}
\end{table*}

\subsection{Effect of the Inter-Sentence Gate}

In order to examine the effectiveness of the proposed inter-sentence gate and inter-sentence information from before-x, we also conducted three additional validation experiments in the test sets: (1) we set all before-x to NULL (the before-x  sentence consists of only stop symbols). (2) we set \(z_t\) to the fixed vector value {\bf 0} for \(NMT_{ISG}\) to block the inter-sentence gate mechanism. (3) we set the context vector \(c_t^a\) of before-x to a random vector, the purpose of which is to test whether the information of the preceding sentence has bad influence on the translation of the current sentence when the preceding sentence is not quite related with the current sentence, for example, a topic change happens between x and before-x. The results are shown in Table 3, from which can find that:

\begin{itemize}
\item When we set before-x as NULL (+NULL in Table 3), the performance  has an obvious decline comparing with that of \(NMT_{ISG}\), but is still better than RNNSearch in term of BLEU score. We conjecture that the reasons for this are twofold. First, in the training process, when before-x dose not exist for the first sentence of a document, we set before-x to a NULL sentence, which makes \(NMT_{ISG}\) model learn the relevant capabilities. That is to say, \(NMT_{ISG}\) treats all sentences in the test sets as the first sentences of documents, where some sentences are correctly handled while others not. Second, the gate mechanism assigns a pretty low rate for NULL before-x during context combination. 

\item When we set the gate weight vector $z$ to a fixed all-zero vector (+$z$={\bf 0} in Table 3), \(NMT_{ISG}\) blocks the inter-sentence information from before-x. From the Table 3, we find a huge loss in performance. The \(NMT_{ISG}\) (+$z$={\bf 0}) is even worse than RNNSearch by 2.31 BLEU points.
Although the preceding sentence information is not used, this \(NMT_{ISG}\) (+$z$={\bf 0}) is not exactly the baseline RNNSearch. There are two groups of parameters in \(NMT_{ISG}\): one group of parameters (new parameters) are related with the inter-sentence gate and the other group of parameters (old parameters) are from RNNSearch but have been optimized towards the maximum usage of inter-sentence information.
When the inter-sentence gate is closed, old parameters are not able to guide the decoder to generate translations that best use the current sentence information. That is the reason why this result is even worse than RNNSearch.

\item When we set the context vector \(c_t^a\) of before-x to a random vector (+RV in Table 3), sampled from a uniform distribution between -1 and 1, the performance is worse than the baseline RNNSearch by 0.46 BLEU points. Setting the \(c_t^a\) to a random vector, the information from the pseudo preceding sentence becomes meaningless, and even has a bad or uncorrelated impact on the translation of the current sentence. However, the drop of the performance is not as big as that of \(NMT_{ISG}\) (+$z$={\bf 0}). This suggests that the gate mechanism is able to effectively shield these useless and counteractive information from a pseudo and random preceding sentence.
\end{itemize}

The three experiments further demonstrate that the proposed inter-sentence gate is able to detect useful information for translation and block unrelated information and reconfirm that inter-sentence information is useful for translation.

\subsection{Analysis on Inter-Sentence Attention}

Many studies \cite{bahdanau2015neural,luong2015effective} on attention-based NMT have proved that attention networks are able to detect alignments between parallel sentence pairs. In our \(NMT_{ISG}\) model, we use two attention networks: the first is built for the correspondences between the current source sentence and its target translation and the second for the correspondences between the preceding sentence and the target translation of the current sentence. 
We are interested in what correspondences the second attention network detect.

We use the entropy as the evaluation criterion to measure how attention weights distribute over words in the preceding sentence before-x for a target word in the translation of the current sentence x. If the attention distribution is even, the entropy will be large. Otherwise, the entropy is small, it suggests that the attention distribution is uneven and that the decoder pays attention to one or several particular words in the preceding sentence when generating a target word for the current sentence. The entropy is computed as follows:
\begin{equation}
H = -\sum_{j=1}^{n} \alpha_jlog\alpha_j
\end{equation}

\begin{figure}[!t]
\centering
\includegraphics[height=1.8in,width=2.7in]{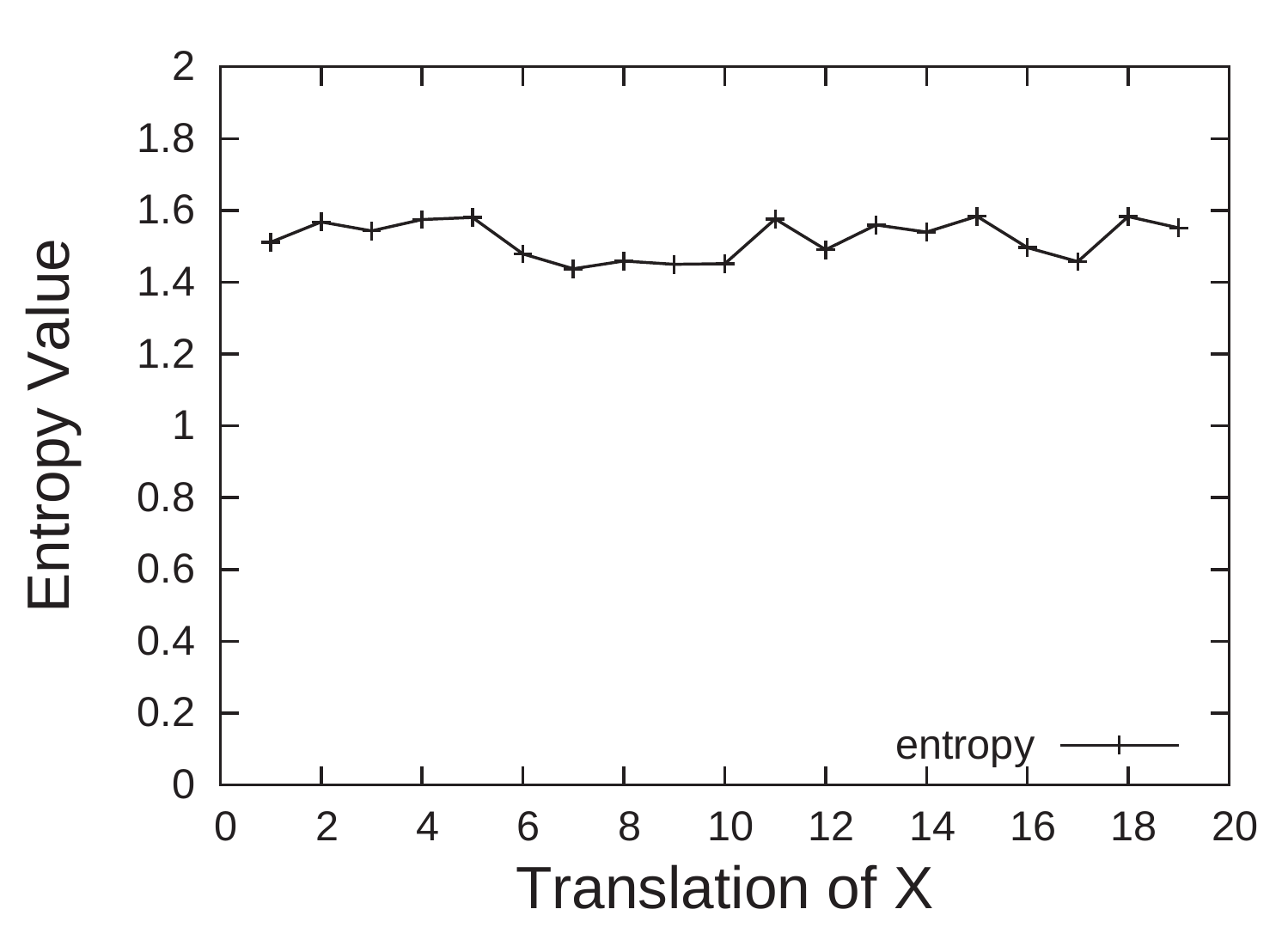}
\caption{The entropy curve for a NULL preceding sentence. This is a sample generated by \(NMT_{ISG}\) in one test set. The horizontal axis represents the translation of a current sentence x. The vertical axis represents the entropy of the preceding sentence before-x obtained from the inter-sentence gate model when translating x.} 
\label{fig:4}
\end{figure}

We calculated entropy values in two cases:
\begin{itemize}
\item The before-x is a NULL sentence, being composed of stop symbols. 
\item The before-x is an ordinary sentence.
\end{itemize}

If before-x is NULL, it cannot provide useful information except for the indicator of being the first sentence of a document. 
The attention weight distribution over stop symbols in before-x when generating a target word is supposed to be uniform.
Figure 3 exactly visualizes such a case when translating the first sentence in a document.
The entropy value in Figure 3 ranges from 1.44 to 1.58 while the entire curve is quite smooth.

Figure 4 demonstrates the second case with an example (in Table 4) where some words  in the preceding sentence reoccur in the current sentence.
When the \(NMT_{ISG}\) model generates the target translation for ``非洲国家(feizhou guojia)'' and ``非洲(feizhou)'', 
the entropy significantly drops as these words have occurred both in the preceding and current sentence. This indicates that the attention network in \(NMT_{ISG}\) successfully captures this word repetition (the most common lexical cohesion device) and convey such an inter-sentence relation collectively with the proposed inter-sentence gate to the prediction of target words via hidden states of the decoder. 

\begin{figure}[!t]
\centering
\includegraphics[height=1.8in,width=2.7in]{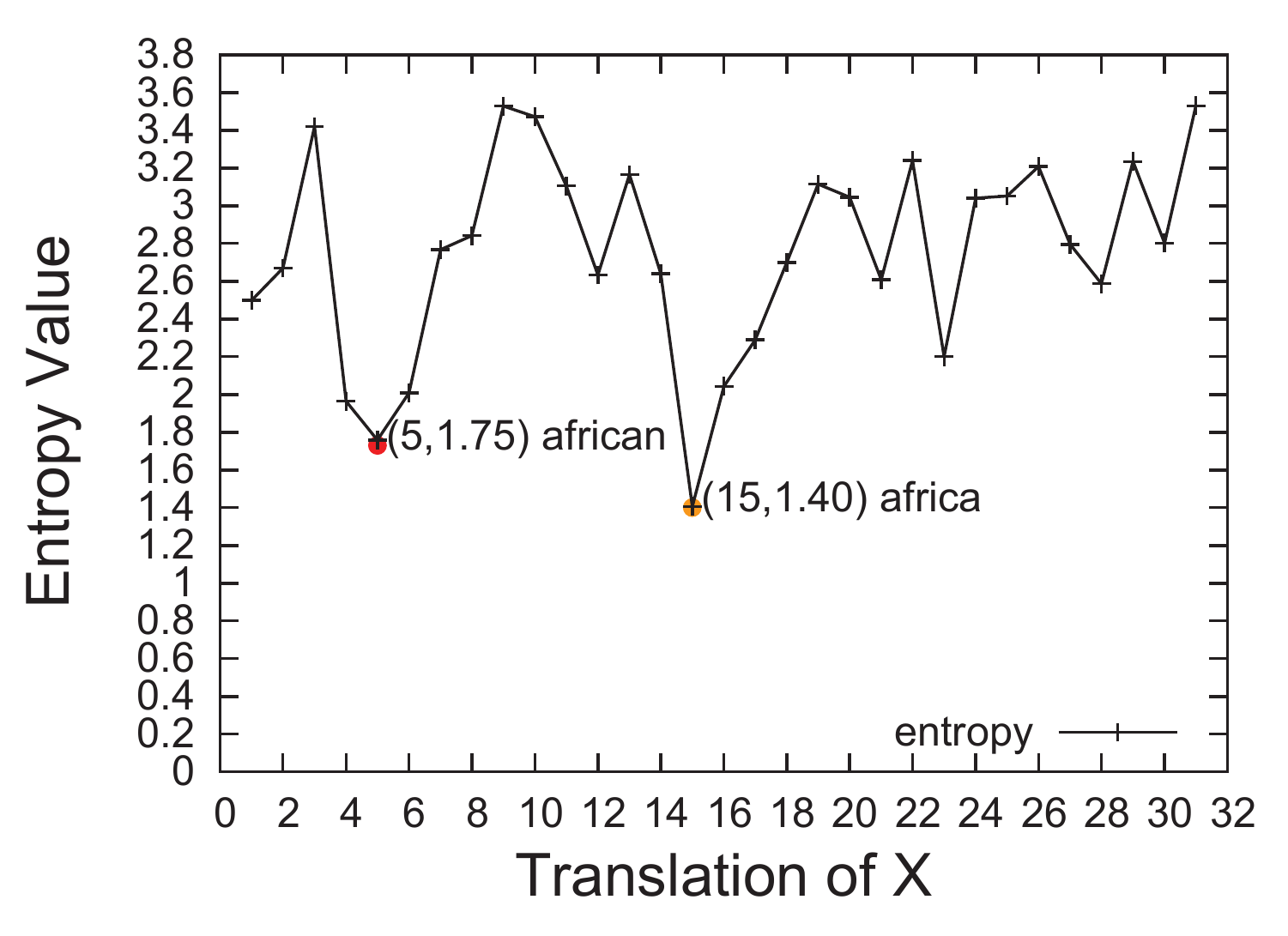}
\caption{The entropy curve for a normal preceding sentence.} 
\label{fig:5}
\end{figure}

\begin{adjustwidth}{-1cm}{-1cm}
\begin{table*}[!t]
\centering
\begin{tabular}{|p{.12\textwidth}| p{.75\textwidth}|}
\hline 
Before-x  & 进入 新 世纪 后 ， 经济 全球化 成为 \textcolor{red}{非洲 国家}面临 的 一个 重大 而 严峻 的 挑战 。 \\
\hline
X         &\textcolor{red}{非洲  国家}领导人 越来越 清楚 地 认识 到,\textcolor{red}{非洲}再也 不 能 坐失良机 , 应 奋起 寻求 应对 之 策 。\\
\hline 
\(NMT_{ISG}\) & the leaders of \textcolor{red}{the african countries} have become more and more clearly aware that \textcolor{red}{africa} can no longer be able to UNK and take the initiative to pursue countermeasures. \\ 
\hline
\end{tabular}
\caption{\label{sample} Example translation genereted by \(NMT_{ISG}\), before-x is the preceding sentence of x. }
\end{table*}
\end{adjustwidth}

\subsection{Analysis on Translation Coherence}

We want to further study how the proposed inter-sentence gate model influence coherence in document translation. For this, we follow \newcite{Lapata2005Automatic} to measure coherence as sentence similarity. First, each sentence is represented by the mean of the distributed vectors of its words. Second, the similarity between two sentences is determined by the cosine of their means.
\begin{equation}
sim(S_1,S_2) = cos(\mu(\vec{S_1}),\mu(\vec{S_2}))\\
\end{equation}
where \(\mu(\vec{S_i})=\frac{1}{|S_i|}\sum_{\vec{w} \in S_i}\vec{w}\), and \(\vec{w}\) is the vector for word \(w\).

We use Word2Vec\footnote{https://code.google.com/p/word2vec/} to obtain the distributed vectors of words and English Gigaword fourth Edition as training data to train Word2Vec. We consider that embeddings from word2vec trained on large monolingual corpus can well encode semantic information of words. We set the vectors of words to 400. 

Table 5 shows the average cosine similarity of adjacent sentences in test sets. From the table, we can find that the \(NMT_{ISG}\) model produces better coherence in document translation than RNNSearch in term of cosine similarity. 

In order to better reflect the performance of the model about coherence, we also provide two examples displayed in Table 6 to verify the impact of inter-sentence information on document-level NMT. In the first example, source x does not have enough context information to correctly translate word ``动作(dongzuo)''. Fortunately, the before-x provides extra information: the background is about politices, which guide the decoder to select a better translation ``action'' for ``动作(dongzuo)''. In the second example, RNNSearch generates different translations for ``驻韩(zhuhan)''  and ``驻南韩(zhunanhan)''. \(NMT_{ISG}\) generates consistent translations for these two different words but with the same meaning.

\begin{table*}[!t]					
\centering
\begin{tabular}{c|cccccc}
\hline 
\bf Model & \bf NIST02 & \bf NIST03 & \bf NIST04 & \bf NIST05  & \bf NIST06 & \bf NIST08 \\ \hline
\hline
RNNSearch            & 0.4536 & 0.4510 & 0.4761 & 0.4677 & 0.3982 & 0.3716 \\
\hline
\(NMT_{ISG}\)        & 0.4744 & 0.4656 & 0.4849 & 0.4816 & 0.4072 & 0.3825 \\
\hline
Human Reference         & 0.5090 & 0.4367 & 0.5100 & 0.5073 & 0.3804 & 0.3911 \\
\hline
\end{tabular}
\caption{\label{font-table7} The average cosine similarity of adjacent sentences in test sets.}
\end{table*}

\begin{adjustwidth}{-1cm}{-1cm}
\begin{table*}[!t]
\centering
\begin{tabular}{|p{.12\textwidth}| p{.77\textwidth}|}
\hline
before-x  & 自从 上个月 巴勒斯坦 强人 阿拉法特 去 逝 后 , 国际 社会 重新 继续 恢复 中东 和平 政策 的 推动 , 以期 早日 结束 以 巴 之间 多年 的 流血 冲突 。\\
\hline
x  & 中东 新闻社 说 , 官员 预测 「 准备 工作 将 会 进行 到 七月 , 然后 再 展开 政治\textcolor{red}{动作} 」\\
\hline
RNNSearch &the UNK news agency said that the officials forecast that ``preparations will be made in july and then political \textcolor{red}{moves} will be taken again.''\\ 
\hline
\(NMT_{ISG}\) & the middle east news agency said, the officials forecast that ``preparations will be made in july and then another political \textcolor{red}{action} will be taken.'' \\
\hline
\hline
before-x  & 美军 第八 军团 司令 康贝尔 中 将 发表 声明 , 此 一 冻结 调防 军令 旨在 确保 \textcolor{red}{驻 南 韩}美军 实力 。\\
\hline
x  & 根据 南韩 与 美国 签订 的 协 防 条约 , 目前 驻 南 韩 美军 人数 约 三万七千 人 , 自 去年 十二月 北 韩 发展 核子 计 画 野心 曝光 以来 , \textcolor{red}{驻 韩}美军 一直 保持 警戒 。\\
\hline
RNNSearch &according to the UNK treaty signed between south korea and the united states, the number of us troops in south korea is about UNK, and the us troops stationed in the \textcolor{red}{rok} since december last year have been kept alert.\\ 
\hline
\(NMT_{ISG}\) & according to the UNK treaty signed between south korea and the united states, the number of us troops stationed in south korea is about UNK. the us troops stationed in \textcolor{red}{south korea} have been maintaining vigilance since last december last year when north korea's nuclear plan was exposed. \\
\hline
\end{tabular}
\caption{\label{sample2} Example translations genereted by \(NMT_{ISG}\).}
\end{table*}
\end{adjustwidth}

\section{Conclusion and Future Work}

In this paper, we have presented a novel inter-sentence gate model for NMT to deal with document-level translation. Experimental results show that the \(NMT_{ISG}\) model achieves consistent and significant improvements in translation quality over strong NMT baselines. In-depth analyses further demonstrate that the proposed model inter-sentence gate is able to capture cross-sentence dependencies and lexical cohesion devices.

The proposed inter-sentence gate model only uses source-side information to capture document-level information for translation. In the future, we would like to integrate target-side information into document-level NMT.

\section*{Acknowledgments}
The present research was supported by the
National Natural Science Foundation of China
(Grant No. 61622209). We would like to thank three anonymous reviewers
for their insightful comments.

\bibliography{colingref}
\bibliographystyle{acl}

\end{CJK}
\end{document}